\pdfoutput=1
\documentclass[11pt]{article}

\usepackage[preprint]{acl}

\usepackage{times}
\usepackage{latexsym}
\usepackage{multirow}
\usepackage{amsmath}
\usepackage{amssymb}

\usepackage[T1]{fontenc}
\usepackage[utf8]{inputenc}

\usepackage{microtype}

\usepackage{inconsolata}

\usepackage{graphicx}

\title{Estimating Text Temperature with Language Models}

\author{Nikolay Mikhaylovskiy \\
NTR Labs / Moscow, Russia \\
Higher IT School of Tomsk State University / Tomsk, Russia \\
\texttt{nickm@ntr.ai} \\}

\begin{document}
\maketitle
\begin{abstract}
Autoregressive language models typically use temperature parameter at inference to shape the probability distribution and control the randomness of the text generated. After the text was generated, this parameter can be estimated using maximum likelihood approach. Following it, we propose a procedure to estimate the temperature of any text, including ones written by humans, with respect to a given language model \footnote{Code is available at https://github.com/nickm197/estimating-text-temperature}. We evaluate the temperature estimation capability of a wide selection of small-to-medium Large Language Models (LLMs). We then use the best-performing Qwen3 14B to estimate temperatures of popular corpora,
finding that while most measured temperatures are close to 1, notable exceptions include Jokes, GSM8K, and AG News (1.1), and Python code (0.9).

\end{abstract}

\section{Introduction}

Recent research (\citet{Bahamondes}, \citet{nakaishi2024criticalphasetransitionlarge}, \citet{mikhaylovskiystates}, \citet{mikhaylovskiy-2025-zipfs}) has shown that LLM-generated text can be considered as a physical system that can undergo phase transitions as the generation temperature changes. This marks the emergence of the Statistical Physics of Language  field. For human-written texts, the temperature is not known a priori. We propose a procedure to estimate the temperature of any text, including ones written by humans, with respect to a given LLM. This  empowers the application of the statistical physics apparatus to human-written texts and provides the essential tooling for this new avenue of research.

\subsection{Autoregressive Probabilistic Language Models}

Probabilistic language models consider sequences

\begin{equation}
t^{(1:m)}= \{t^{(1)},t^{(2)},…,t^{(m)} \}
\end{equation}

of tokens from the lexicon $L$. An autoregressive language model estimates the probability of such a sequence
\begin{gather}
\label{eq1}
\begin{split}
P(t^{(1:m)})=
P(t^{(1)})P(t^{(2)}|t^{(1)})…\\
P(t^{(m)}|t^{(1:m-1)})   \\
= \prod_{k=1}^{m}{P(t^{(k)}|t^{(1:k-1)}})
\end{split}
\end{gather}
using the chain rule.

\subsection{Text Generation with a Language Model}
 
Given an input text as a context, the goal of open-ended generation is to produce a coherent continuation of the text \citep{Holtzman2020The}. More formally, given a sequence of $m$ tokens $t^{(1)} \dots t^{(m)}$ as a context, the objective is to generate the next $n$ continuation tokens, resulting in the completed sequence $t^{(1)} \dots t^{(m+n)}$. This is achieved through the use of the left-to-right text probability decomposition, which is used to generate the sequence one token at a time, using a particular decoding strategy.

A common approach to text generation is to shape a probability distribution through temperature \citep{ACKLEY1985147}. 
Let 
\begin{itemize}
\item $u^{(i)}$ be the vector of logits at the step $i$, indexed with $l \in L$. 
\item $u_{obs}^{(i)}$ be the specific logit value corresponding to the token $t^{(i)}$. 
\end{itemize}

The probability of observing the specific token at step $i$ is then re-estimated as:

\begin{equation}
\label{eq3}
p(t^{(i)}|u^{(i)}, T) = \frac{\exp(u_{obs}^{(i)}/T)}{\sum_{l} \exp(u_{l}^{(i)}/T)}
\end{equation}

Setting $T \in [0, 1)$ skews the distribution towards high-probability events, and, similarly, $T \in (1, \infty)$ skews the distribution towards low-probability events.

\section{Estimating Temperatures}

\subsection{Estimating the Temperature of a Generated Sequence}
 
Suppose that we have logits for the multiple generation steps and the actual generated tokens for the same steps. What is the maximum likelihood estimate of temperature $T$? We can treat it as a classic parameter estimation problem.

The total log-likelihood $\mathcal{L}(T)$ for all $N$ steps is the sum of the individual log-probabilities:

\begin{equation}	
\mathcal{L}(T) = \sum_{i=1}^{N} \log \left( \frac{\exp(u_{obs}^{(i)}/T)}{\sum_{l} \exp(u_{l}^{(i)}/T)} \right)
\end{equation}

To find the MLE, we take the derivative of $\mathcal{L}(T)$ with respect to $T$ and set it to zero. 

\begin{gather}
\label{eq2}
\begin{split}	
\frac{d\mathcal{L}}{dT} = \\
\sum_{i=1}^{N} \left[ -\frac{u_{obs}^{(i)}}{T^2} - \left( -\frac{1}{T^2} \sum_{l} u_{l}^{(i)} p(l|u^{(i)}, T) \right) \right] \\
 = \frac{1}{T^2} \sum_{i=1}^{N} \left[ \mathbb{E}[u^{(i)}|T] - u_{obs}^{(i)} \right]
\end{split}
\end{gather}

Setting the derivative (\ref{eq2}) to zero, we get for the MLE $\hat{T}$:

\begin{equation}
\label{MLECondition}
\sum_{i=1}^{N} u_{obs}^{(i)} = \sum_{i=1}^{N} \mathbb{E}[u^{(i)}|\hat{T}],
\end{equation}

where
\begin{equation}
\label{eq7}
\mathbb{E}[u^{(i)}|\hat{T}] = \sum_{l} u_{l}^{(i)} \frac{\exp(u_{l}^{(i)}/\hat{T})}{\sum_{k} \exp(u_{k}^{(i)}/\hat{T})}
\end{equation}

Thus, the maximum likelihood estimate for $T$ is the temperature at which the sum of the observed logits equals the sum of the expected logits predicted by the model.

\subsection{Estimating the Temperature of Any Text}

Now, observe that the text we work on in the previous subsection does not need to be a generated one. As soon as we have tokenized a text and have estimated the probability distributions for all the tokens in the text with some probabilistic model (for example, an LLM), we can estimate the text temperature using (\ref{MLECondition}). This estimate will, of course, depend on the LLM model used. 

To specifically find the temperature, we numerically solve equation (\ref{MLECondition}) for $T$, given the logits for all the tokens and specific tokens present in the text.

%\subsection{Text Temperature and Perplexity}
%
%The classical metric of interplay between a probabilistic language model and a text is perplexity, introduced by \citet{10.1121/1.2016299}. Perplexity is defined as the exponential of the negative log-likelihood. Therefore, maximizing likelihood is mathematically identical to minimizing perplexity.
%\begin{equation}	
%\hat{T} = \underset{T}{\operatorname{argmin}} \big( \text{Perplexity}%(text | \text{Model}, T) \big)
%\end{equation}

%Normally, perplexity is reported at $T=1$. Calculating perplexity at the optimal $T=\hat{T}$ measures the best possible fit the model can achieve just by scaling its confidence.

\section{Estimating Temperature of LLM-Generated Texts}

\begin{figure}[t]
  \includegraphics[width=1.0\columnwidth ]{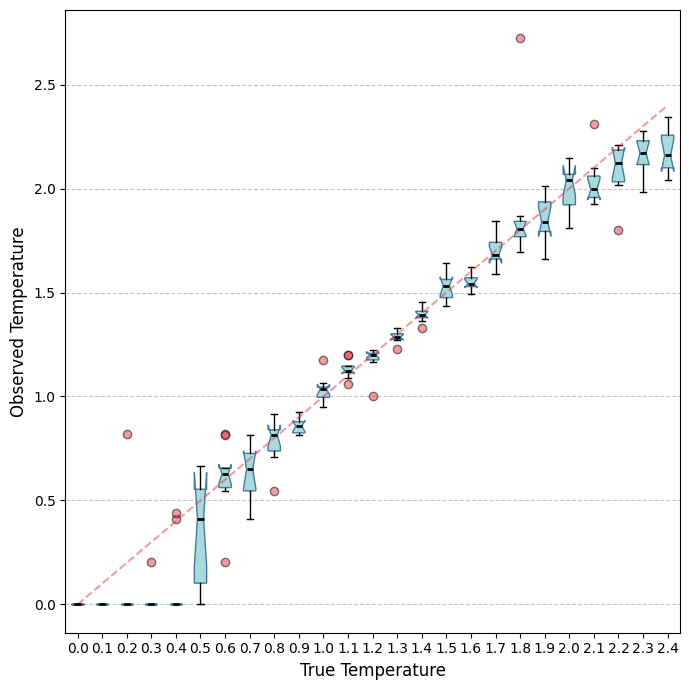}
  \caption{Estimated vs. generation temperature for granite-4.0-micro.}
   \label{fig1}
\end{figure}
\begin{figure}[t]
  \includegraphics[width=1.0\columnwidth ]{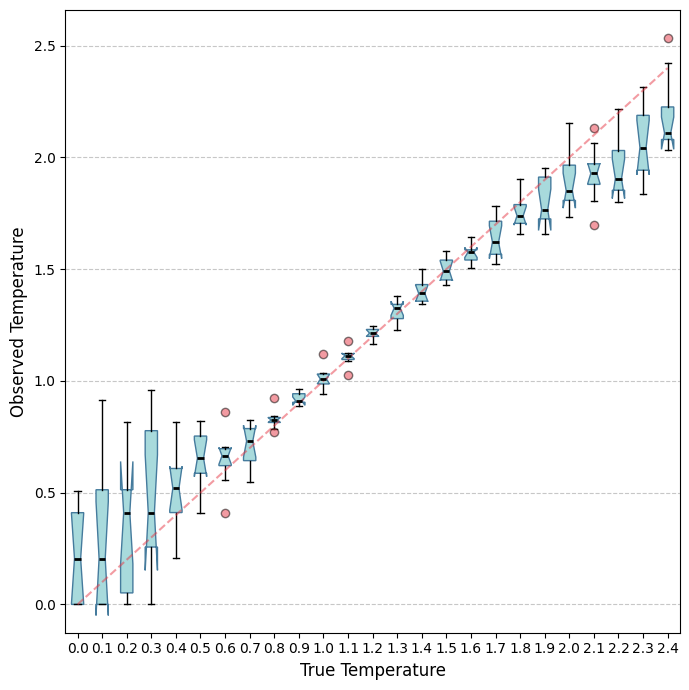}
  \caption{Estimated vs. generation temperature for Llama-3.2-3B.}
   \label{fig2}
\end{figure}

To better understand the properties of the proposed temperature estimation approach, we generate texts with an assortment of LLMs in small to medium sizes representing different LLM families, finetuning levels and architectures:

\begin{itemize}
\item Qwen3 0.6B,1.7B 4B, 8B, 14B \cite{yang2025qwen3technicalreport} (normal, base and 4B-Thinking-2507) 
\item Phi-4-mini \cite{microsoft2025phi4minitechnicalreportcompact} (reasoning and instruct) 
\item Llama 3.1 8B, 3.2 1B, 3B, 11B-Vision \cite{grattafiori2024llama3herdmodels} (base and instruct)
\item gemma-3 270m, 1b, 4b, 12b \cite{gemmateam2025gemma3technicalreport}
\item DeepSeek-R1 distills: 0528-Qwen3-8B, Qwen-1.5B, Qwen-7B, Qwen-14B, and Llama-8B \cite{deepseekai2025deepseekr1incentivizingreasoningcapability}
\item granite-4.0-micro \cite{granite2025} (base and normal) 
\end{itemize}

We use 4-bit BitsAndBytes NF4 quantized models through the HuggingFace Transformers library \citep{wolf-etal-2020-transformers}. All the texts are generated from a single random seed-controlled token using batch inference. We generate 10 texts 200 tokens long for each temperature in a range from $0.001$ to $2.401$ with step $0.1$. We do not use top-k, top-p or any other decoding parameters such as no-repeat. To estimate the temperature, we use SciPy \cite{2020SciPy-NMeth} root finding function \verb|root_scalar| for the reverse temperature with bracket $[10^{-2}, 10^4]$ and initial value $5*10^3$. To speed up the computations we compute softmax (\ref{eq3}, \ref{eq7}) using PyTorch on GPU. The code was mostly written by Gemini.

\subsection{What Does the Measured Temperature Say About the Generation Temperature?}

\label{sec:temp2temp}

In the first series of experiments, we estimate the temperature of texts generated by an LLM with the same LLM. For most LLMs, the estimated temperature insignificantly differs from the generation temperature. 

For all the models, we observe pronounced and systematic differences between estimated and generation temperatures in areas of low and high temperatures (see Figures~\ref{fig1}, \ref{fig2}). The difference in the low temperature area is the result of saturation when the generation with different temperatures produces one and the same result. Thus, the temperature estimation at low temperatures is an ill-posed problem that can be resolved using some kind of regularization yet to be found. This constitutes an exciting area for future research. We don't yet have an explanation for the discrepancy between estimated and generation temperatures at high temperatures. 

\subsection{What Does the Measured Temperature Say About the Model?}

\label{sec:temp2model}

\begin{figure}[t!]
  \includegraphics[width=1.0\columnwidth ]{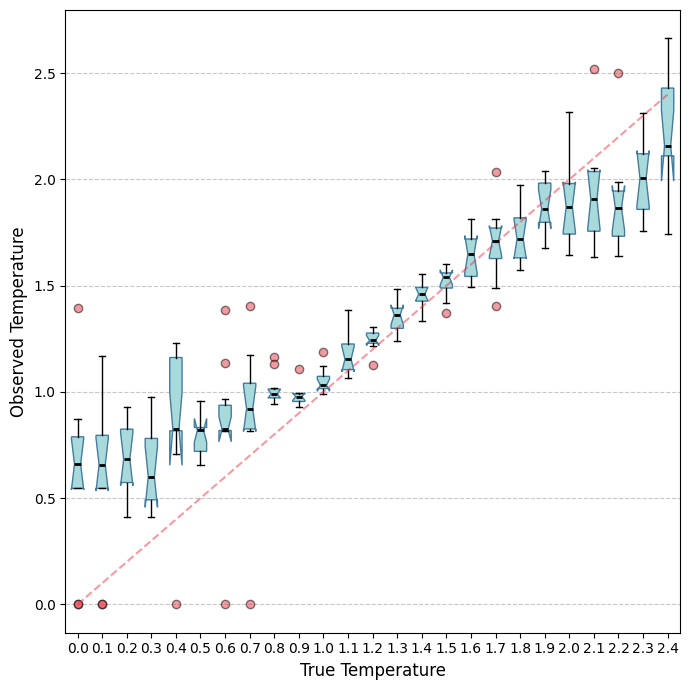}
  \caption{Qwen3-8B estimates Qwen3-4B temperature}
   \label{fig3}
\end{figure}

\begin{figure}[t!]
  \includegraphics[width=1.0\columnwidth ]{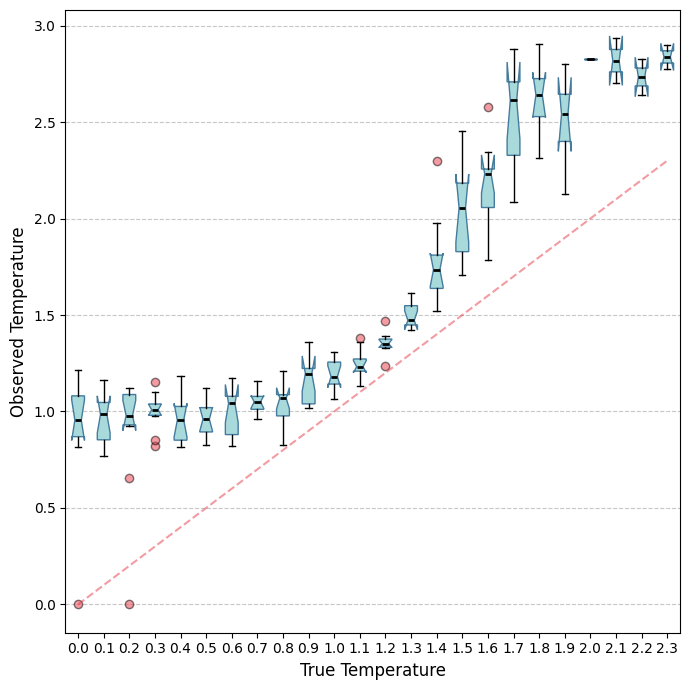}
  \caption{DeepSeek-R1-0528-Qwen3-8B estimates DeepSeek-R1-Distill-Qwen-1.5B temperature}
   \label{fig4}
\end{figure}

\begin{figure}[t]
  \includegraphics[width=1.0\columnwidth ]{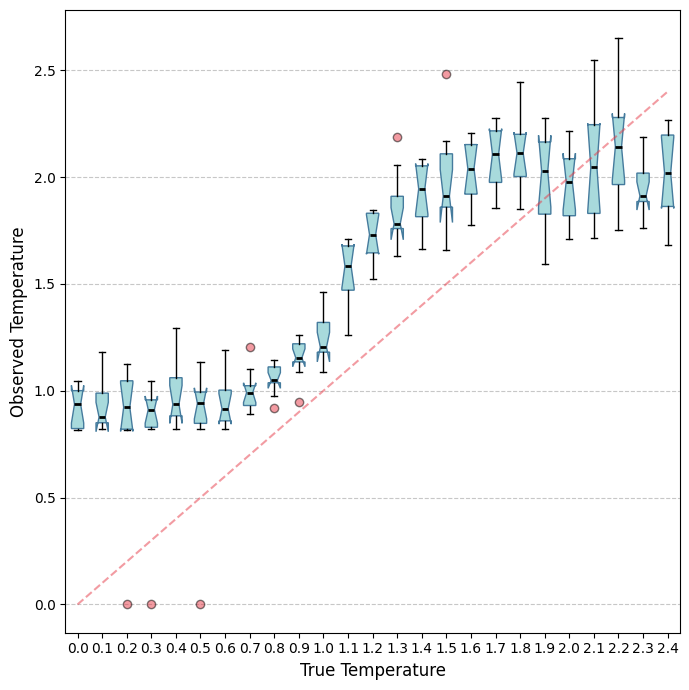}
  \caption{DeepSeek-R1-0528-Qwen3-8B estimates Meta-Llama-3.1-8B-Instruct temperature}
   \label{fig5}
\end{figure}

\begin{figure}[t]
  \includegraphics[width=1.0\columnwidth ]{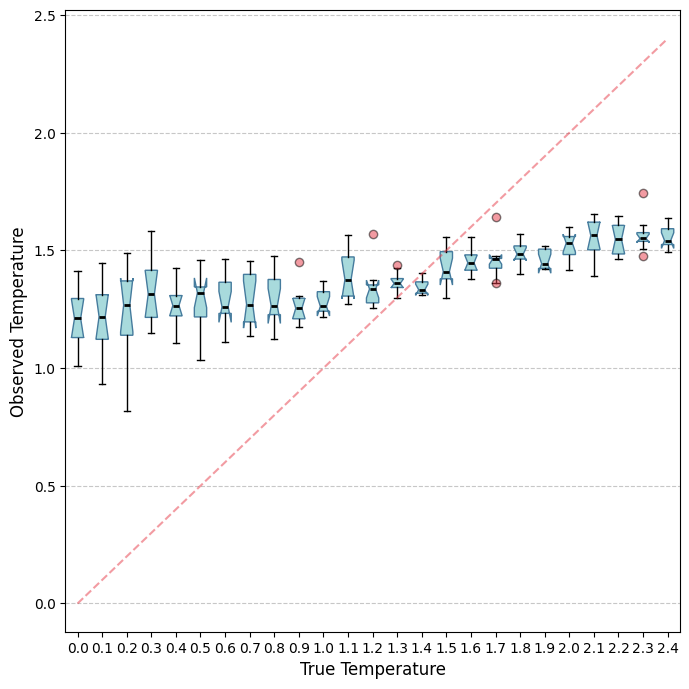}
  \caption{gemma-3-1b-it estimates Qwen3-14B temperature}
   \label{fig6}
\end{figure}

In the second series of experiments, we estimate the temperature with an LLM different from the one used to generate the text. The observed behaviors can be put into several qualitatively distinct groups. In one group, the graph of estimated vs. true temperature does not differ much from what a graph would be for the same LLM. This happens mostly for similar models in the families of Qwen, granite and LLama (see Figure~\ref{fig3} for an example). 

The other group consistently overestimates the generation temperature 
(see Figure~\ref{fig4} for an example). This is typical for DeepSeek distills and happens in some other cases, but never for base models. We attribute this to narrower probability distributions of finetuned, especially reasoning, models.

Yet another group consists of model pairs that produce a pronounced S-shaped graph (see Figure~\ref{fig5}). Finally, there are model pairs that show little to no correlation between the generation and the estimated temperatures (see Figure~\ref{fig6} for an example). This is often the case when either generator or estimator models are gemma of Phi. 

\begin{figure}[t!]
  \includegraphics[width=1.0\columnwidth ]{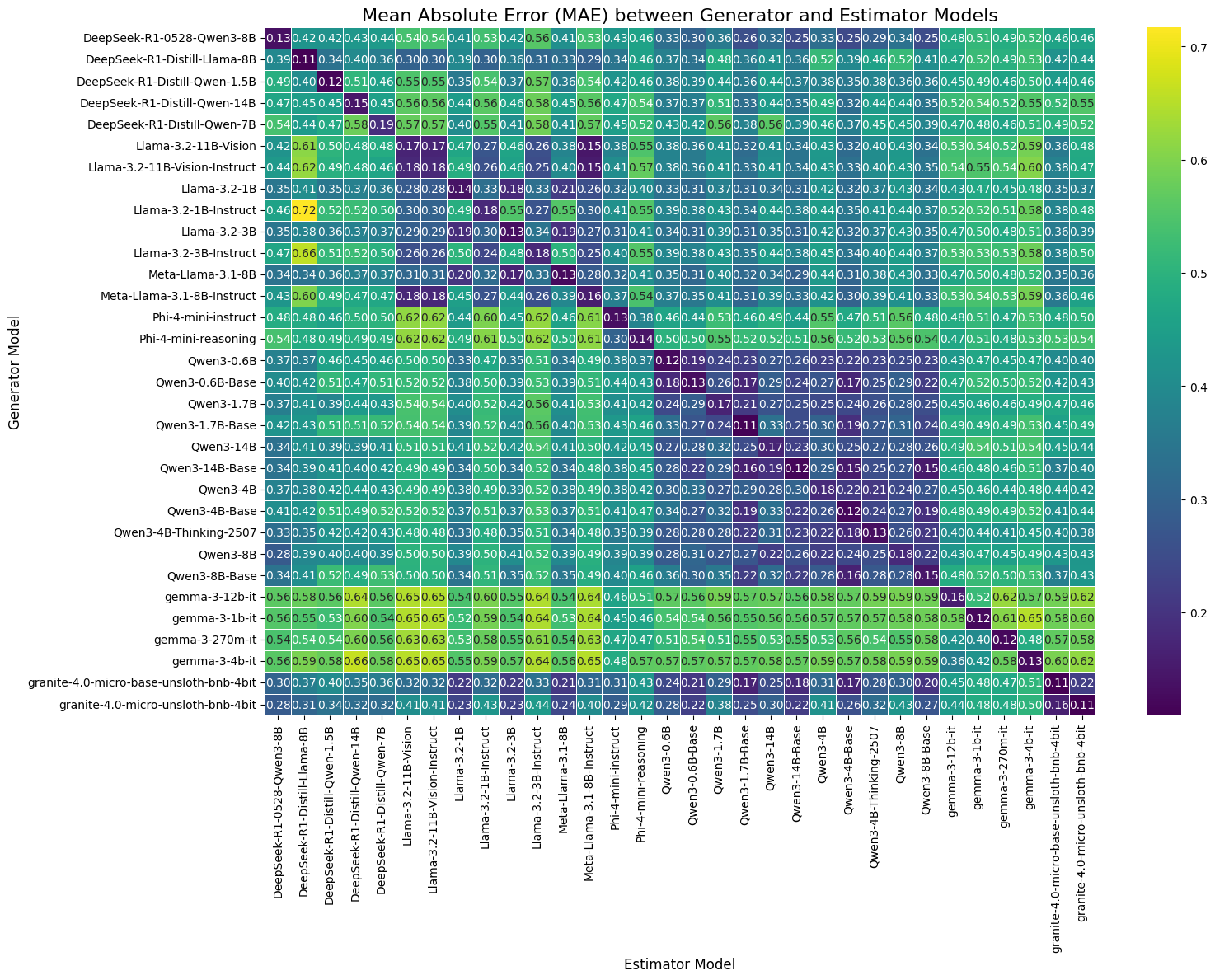}
  \caption{Heatmap of MAE between models}
   \label{fig7}
\end{figure}

To digitize these observations, we calculated statistical metrics of goodness such as MAE, $R^2$, Pearson $p$ between observed and generation temperatures for each pair of models. Figure~\ref{fig7} presents a heatmap of MAE between the models. From this heatmap we can conclude that:
\begin{itemize}
\item Models in Qwen, LLama and granite families reasonably well estimate temperature of models from these families. They estimate  even better when the generator and estimator models are from the same family. 
\item DeepSeek, gemma and Phi models are not good at both estimating and being estimated 
\item Base models are overall slightly better temperature estimators than finetuned ones. Base models better estimate base models and finetuned better estimate finetuned ones. This is especially pronounced for Llama family.
\item Larger models are slightly better estimators 
\item Qwen models are the best estimators, the absolute leader being Qwen3-14B-base
\end{itemize}

\section{Estimating Temperature of Human-written Corpora}

\begin{table}
\centering
\begin{tabular}{| l | l | l |}
\hline
Dataset & Mean $t$ & std \\
\hline
WikiText  & 1.0585  & 0.0645  \\
\hline
Poetry  & 1.0089  & 0.1044  \\
\hline
Jokes  & 1.1003  & 0.1037  \\
\hline
GSM8K  & 1.0942  & 0.0992  \\
\hline
Code/Python  & 0.9242  & 0.1539  \\
\hline
IMDB  & 1.0244  & 0.0385  \\
\hline
HH RLHF  & 1.0063  & 0.0641  \\
\hline
AG News  & 1.1023  & 0.0734  \\
\hline
Yelp  & 1.0349  & 0.0545  \\
\hline

\end{tabular}
\caption{Temperatures of corpora}
\label{t1}
\end{table}

Based on the results in the previous section, we use Qwen3 14B Base as the preferred LLM for the temperature estimation. We estimate temperature of several diverse datasets: WikiText-103 \citep{merity2016pointer}, Poetry dataset \citep{mervepoetry}, Jokes dataset \citep{ysharmashort_jokes}, GSM8K \citep{cobbe2021gsm8k}, Python code \citep{python_code_instructions_18k_alpaca}, IMDB \citep{maas-EtAl:2011:ACL-HLT2011}, HH RLHF \citep{bai2022traininghelpfulharmlessassistant, ganguli2022redteaminglanguagemodels}, AG News \citep{10.5555/2969239.2969312, 10.1145/1062745.1062778, Corso2005RankingAS}, Yelp \citep{10.5555/2969239.2969312}.  We sample 300 texts from each dataset and average temperatures over texts in a dataset. The results are in the Table~\ref{t1}.

\section{Conclusion}

We proposed a procedure to estimate the temperature of any text, with respect to a given language model. We evaluated the temperature estimation capability of a wide selection of small-to-medium LLMs. 

The estimated temperatures are all close to 1. $T \approx 1$ is an empirical validation of the Language Modeling objective. Since models are trained to minimize cross-entropy on human corpora, a perfectly calibrated model should perceive human text as having $T=1$. However, the deviations are instructive. The statistically significant shift for Jokes ($1.1$) vs Code ($0.9$)  provides a quantitative measure of the "entropy requirement" for different tasks. This confirms that "creativity" (jokes) requires a flatter distribution (higher $T$) while "syntax" (code) requires a sharper one (lower $T$).

%\subsection{Potential Applications} 
%Some of the potential applications of our results are:

%\begin{itemize}
%\item Domain-Adaptive Decoding. Our findings (Table~\ref{t1}) show distinct "intrinsic temperatures" for domains (Python, Jokes). This enables automatic parameter selection: an agent can estimate the temperature of the prompt’s domain and adjust the generation temperature dynamically, rather than relying on a static user setting.

%\item Model Forensics and Detection. Text claiming to be Python code but estimating at $T=1.2$ is likely hallucinated or badly generated.

%\item Quantifying Over-Alignment. The failure of specific models (DeepSeek/Phi) to estimate temperature accurately (Figure~\ref{fig7}) acts as a diagnostic for distributional collapse caused by aggressive RLHF or distillation.
%\end{itemize}

%Of course, each of these potential applications is just a hypothesis that requires further research.

\section*{Acknowledgements}

The idea of this work is due to Boris Orekhov asking me "Can you measure the temperature of a language?". I thank him for asking such questions.

\section*{Limitations}

The experiments were conducted for English only. This may limit generalization to other languages. This is a topic for immediate further research. 

The list of LLM tested is limited to small-to-medium models. Scaling laws, if any, are unclear.

\bibliography{anthology,custom}
% \bibliography{custom}

\end{document}